\documentclass[10pt,twocolumn,letterpaper]{article}

\usepackage{cvpr}
\usepackage{times}
\usepackage{epsfig}
\usepackage{graphicx}
\usepackage{amsmath}
\usepackage{amssymb}

\usepackage{verbatim}
\usepackage{tabularx}
\usepackage{booktabs}
\usepackage{multirow}
\usepackage{algorithm}
\usepackage{algorithmic}

\usepackage[pagebackref=true,breaklinks=true,letterpaper=true,colorlinks,bookmarks=false]{hyperref}

\cvprfinalcopy 

\def\cvprPaperID{1522} 

\ifcvprfinal\fi
\begin{document}

\title{Weakly-Supervised Action Segmentation with \\ Iterative Soft Boundary Assignment}

\author{Li Ding\thanks{Work performed when the author was with University of Rochester.}\\
Massachusetts Institute of Technology\\
{\tt\small liding@mit.edu}
\and
Chenliang Xu\\
University of Rochester\\
{\tt\small chenliang.xu@rochester.edu}
}

\maketitle
\thispagestyle{empty}

\begin{abstract}
In this work, we address the task of weakly-supervised human action segmentation in long, untrimmed videos. 
Recent methods have relied on expensive learning models, such as Recurrent Neural Networks (RNN) and Hidden Markov Models (HMM). However, these methods suffer from expensive computational cost, thus are unable to be deployed in large scale. To overcome the limitations, the keys to our design are efficiency and scalability. We propose a novel action modeling framework, which consists of a new temporal convolutional network, named Temporal Convolutional Feature Pyramid Network (TCFPN), for predicting frame-wise action labels, and a novel training strategy for weakly-supervised sequence modeling, named Iterative Soft Boundary Assignment (ISBA), to align action sequences and update the network in an iterative fashion. The proposed framework is evaluated on two benchmark datasets, Breakfast and Hollywood Extended, with four different evaluation metrics. Extensive experimental results show that our methods achieve competitive or superior performance to state-of-the-art methods.
\end{abstract}


\section{Introduction}

One of the major challenges in video understanding~\cite{KrHaReICCV2017, XuMeYaCVPR2016, TaZhStCVPR2016, YuWaHuCVPR2016, ZhXuCoAAAI2018, XuHsXiCVPR2015, HeWaShICCV2017} is to localize and classify human actions in long, untrimmed videos, of which usually the requirement is to predict per-frame semantic labels. Recently, many approaches~\cite{yeung2015every, kuehne2016end, Lea_2017_CVPR, singh2016multi, ding2017tricornet} have been introduced to address this problem in fully-supervised setting, relying on the densely-annotated video data such as~\cite{kuehne2014language, bojanowski2014weakly}. 
However, such data is usually too tedious to obtain, which makes these methods not scalable under real-world condition. Furthermore, annotating the precise temporal boundaries between actions is also a challenge for humans; the resulting data will be inconsistent and less likely to be relied on for learning in larger scale.


As a result, there is an increasing attention on methods~\cite{Richard_2017_CVPR, kuehne2017weakly, huang2016connectionist, bojanowski2014weakly} that focus on solving the problem under weaker supervision, \eg with \textit{action transcripts}. In this case, an \textit{action transcript} refers to a set of action units organized by their occurrence ordering in a video without precisely locating their temporal boundaries. This kind of labeling is much easier to obtain, and can even be automatically generated from other meta information~\cite{alayrac2016unsupervised, laptev2008learning, marszalek2009actions}. 
Our paper aligns with this set of weakly-supervised works and assumes only access to action transcripts of training. 

\begin{figure}[t]
	\begin{center}
		\includegraphics[width=\linewidth]{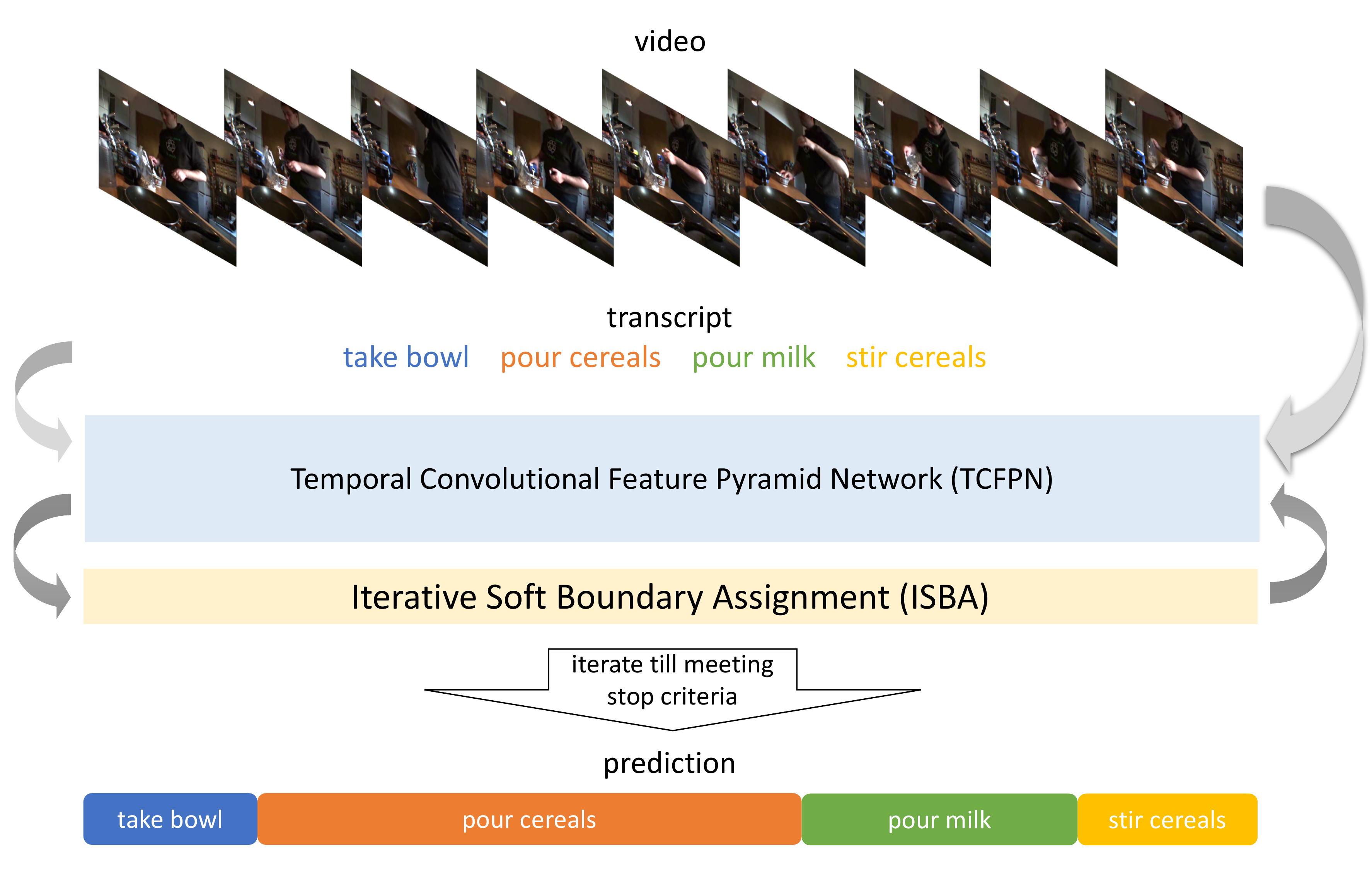}
	\end{center}
    \caption{Overview of the proposed framework with Iterative Soft Boundary Assignment (ISBA) and Temporal Convolutional Feature Pyramid Network(TCFPN).}
    \label{brief}
\end{figure}

However, current methods stick to use Recurrent Neural Networks (RNN) to encode the video data and use different algorithms, such as Extended Connectionist Temporal Classification (ECTC)~\cite{huang2016connectionist} and Hidden Markov Model (HMM)~\cite{Richard_2017_CVPR,kuehne2017weakly}, to go through all the possible action sequences and find the one with maximal likelihood. Both parts, i.e., network and learning model, suffer from expensive computational cost. For example, in \cite{Richard_2017_CVPR}, videos have to be cut into small chunks to enable RNN training. Besides, \cite{Richard_2017_CVPR, kuehne2017weakly} also introduce iterative learning approaches that train until converge, which do not consider the condition of overfitting. 


To overcome the above collective limitations, we propose a novel action segmentation framework for weakly supervised learning, which consists of a new temporal convolutional network, named Temporal Convolutional Feature Pyramid Network (TCFPN), for predicting frame-wise action labels, and a novel training strategy, named Iterative Soft Boundary Assignment (ISBA), to align action sequences and update the network in an iterative fashion (see Fig.~\ref{brief} for an overview). Both parts feature efficient and highly parallelizable computation, e.g., not using any recurrency or Markovian process. We also propose a specific stop criteria that can effectively evaluate the current training process and prevent overfitting. 

Concretely, the ISBA allows us to iteratively train a temporal segmentation network with the training target generated from action transcript, and refine the action transcript based on the inference of current network. Each time, the proposed ISBA looks at the boundary frames between two actions and decides an update to the action boundary based on the current inference result. This leads to increased efficiency than other methods~\cite{Richard_2017_CVPR,huang2016connectionist}, because unlike these methods that try to find the optimal action sequence with maximal likelihood after going through all the candidates during each iteration, ISBA uses a simple-yet-effective algorithm and tries to form a reasonable probabilistic distribution of different actions. Such process takes the idea from Expectation-Maximization (EM) algorithm~\cite{dempster1977maximum}, but features a more randomized behavior.

We also propose a soft boundary mechanism that puts weaker supervision on the boundary frames, which can further help the network find most discriminative patterns for learning an action model. 
Our proposed stop criteria is based on video-level recognition loss and prevents a network from overfitting to the iterative training in ISBA.


In addition, we propose TCFPN, a convolution-based neural network that achieves state-of-the-art performance on temporal modeling with exceeding speed than RNN-based methods. It has a pyramid structure to make use of both low-level and high-level features, with lateral connection~\cite{Lin_2017_CVPR} to reduce computation cost.



For evaluation, we use two benchmark datasets, Breakfast and Hollywood Extended, both of which are among the largest datasets for evaluating action segmentation. Four different metrics are used to comprehensively evaluate the performance.
Extensive experimental results show that our methods achieve competitive or superior results than state-of-the-art methods on both datasets. 

Our contributions are three-fold. First of all, we propose ISBA training strategy with novel stop criteria for weakly-supervised action segmentation and alignment. The ISBA is independent from a chosen segmentation network. Secondly, we propose TCFPN, a new temporal convolutional network for action segmentation. Thirdly, our whole system combining TCFPN and ISBA achieves state-of-the-art performance on weakly-supervised action segmentation and alignment, with exceeding efficiency and scalability. 





\section{Related Work}

We organize the related work on video action modeling into two sets: fully-supervised methods and weakly-supervised methods, and discuss them next.

\noindent \textbf{Fully-Supervised Methods.} \quad Many existing works in this category use frame-level features as input and then build temporal models on the whole video sequence. Yeung \etal~\cite{yeung2015every} propose an attention LSTM network to model the dependencies of the input frame features in a fixed-length window. Singh \etal~\cite{singh2016multi} present a multi-stream bi-directional recurrent neural network for fine-grained action detection task. Kuehne \etal~\cite{kuehne2016end} introduce an end-to-end generative framework for action segmentation using the HTK system, and focus on the part of feature extraction. Lea \etal~\cite{Lea_2017_CVPR} devise two temporal convolutional networks for action segmentation and detection, transforming successful approaches from speech recognition. Ding \etal~\cite{ding2017tricornet} further introduce a hybrid temporal convolutional and recurrent network that also learns action ordering, but suffers high computation cost.

\noindent \textbf{Weakly Supervised Methods.} \quad A variety of different approaches have been explored for the task of weakly supervised action labeling. Bojanowski \etal~\cite{bojanowski2014weakly} formulate the temporal assignment problem and propose Ordering Constrained Discriminative Clustering (OCDC), with the introduction of Hollywood extended dataset. Huang \etal~\cite{huang2016connectionist} propose ECTC which enforces the action alignments to be consistent with frame-wise visual similarities. Kuehne~\cite{kuehne2017weakly} use HMM to model the action and set the ground truth to be the sequence that maximizes the likelihood of all possible sequences. They iteratively refine the segmentation. Following a similar pipeline, Richard \etal~\cite{Richard_2017_CVPR} propose an iterative fine-to-coarse sub-action modeling mechanism with RNN and HMM. Our method is in this category. Comparing to previous works, our method runs fast and achieves state of the art performance, which we show in Sec.~\ref{sec:exp}.

\section{Temporal Convolutional Feature Pyramid Network (TCFPN)}

\begin{figure}[t]
	\begin{center}
		\includegraphics[width=\linewidth]{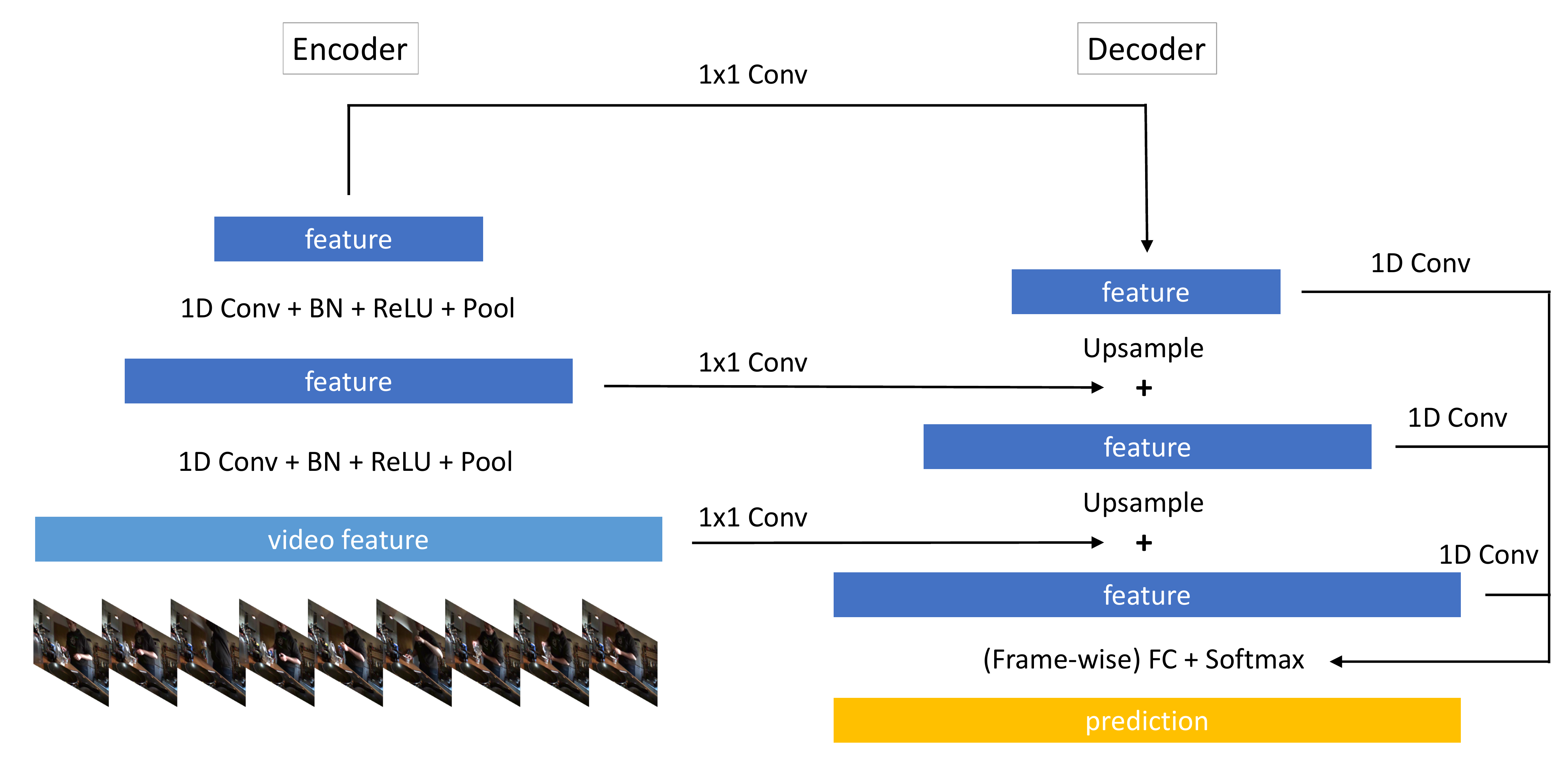}
	\end{center}
	\caption{Structure overview of TCFPN. The proposed network extends the original ED-TCN~\cite{Lea_2017_CVPR} by adding lateral connections~\cite{Lin_2017_CVPR} between encoder and decoder.} 
	\label{tcfpn}
\end{figure}

In order to counter the weakly supervised action segmentation task, a good temporal segmentation network is essential to the whole system. We set the requirements for a \textit{good} segmentation network for our task: (1) it should be able to learn from the coarse ground truth; and (2) it needs to run efficiently enabling the iterative training process. As a result, we adopt the Encoding-Decoding Temporal Convolutional Network (ED-TCN)~\cite{Lea_2017_CVPR} as our baseline and propose an improved structure: TCFPN. We achieve this by adding the lateral connection mechanism proposed in \cite{Lin_2017_CVPR}, which, we note, is for a different task, i.e., object detection. We adapt it here for action segmentation. Figure~\ref{tcfpn} shows an overview of the structure of our proposed network.

TCFPN retains an encoder-decoder architecture. Both encoder and decoder parts consist of $K$ layers of features. We define the encoding layers as $L_{E}^{(i)}$, and the decoding layer as $L_{D}^{(i)}$, for $i = 1,2,\dots,K$, where $K>0$ is the depth parameter that can vary based on the size of dataset. Empirically, we set $K=3$ for all of our experiments. 

In the encoder part, $L_{E}^{(1)}$ is the video feature extracted from each frame. For $i>1$, each layer $L_{E}^{(i)}$ is the output of a combination of operations, \ie, temporal (1D) convolution, batch normalization, ReLU nonlinearity, and max pooling across time. 

In the decoder part, $L_{D}^{(1)}$ is computed from $L_{E}^{(K)}$ with a $1\times1$ convolution with a desired number of filters, which serves mainly as dimension reduction. For $i>1$, each layer $L_{D}^{(i)}$ is computed by the lateral connection with $L_{E}^{(K+1-i)}$, which is the element-wise sum of up-sampled $L_{D}^{(i-1)}$ and the result of applying $1\times1$ convolution on $L_{E}^{(K+1-i)}$. Note that all the $1\times1$ convolution operations in the decoder part have the same number of filters. This is required for element-wise sum in lateral connection and can reduce the dimension of high-level features. The key idea of decoder part is to add high-level semantic information into low-level dense feature maps. 

Finally, each decoder layer $L_{D}^{(i)}$, $i = 1,2,\dots,K$ is operated by another temporal convolution to reduce the aliasing effect of upsampling. A frame-wise fully-connected layer with softmax activation is used to output the class probabilities at each time step for all $K$ layers. The final prediction is averaged through all these layers. Such design naturally combines coarse, semantically-strong features with fine, semantically-weak features in a pyramidal hierarchy, with little extra computation expense.


\section{Iterative Soft Boundary Assignment (ISBA)}

\begin{figure}[t]
	\begin{center}
		\includegraphics[width=\linewidth]{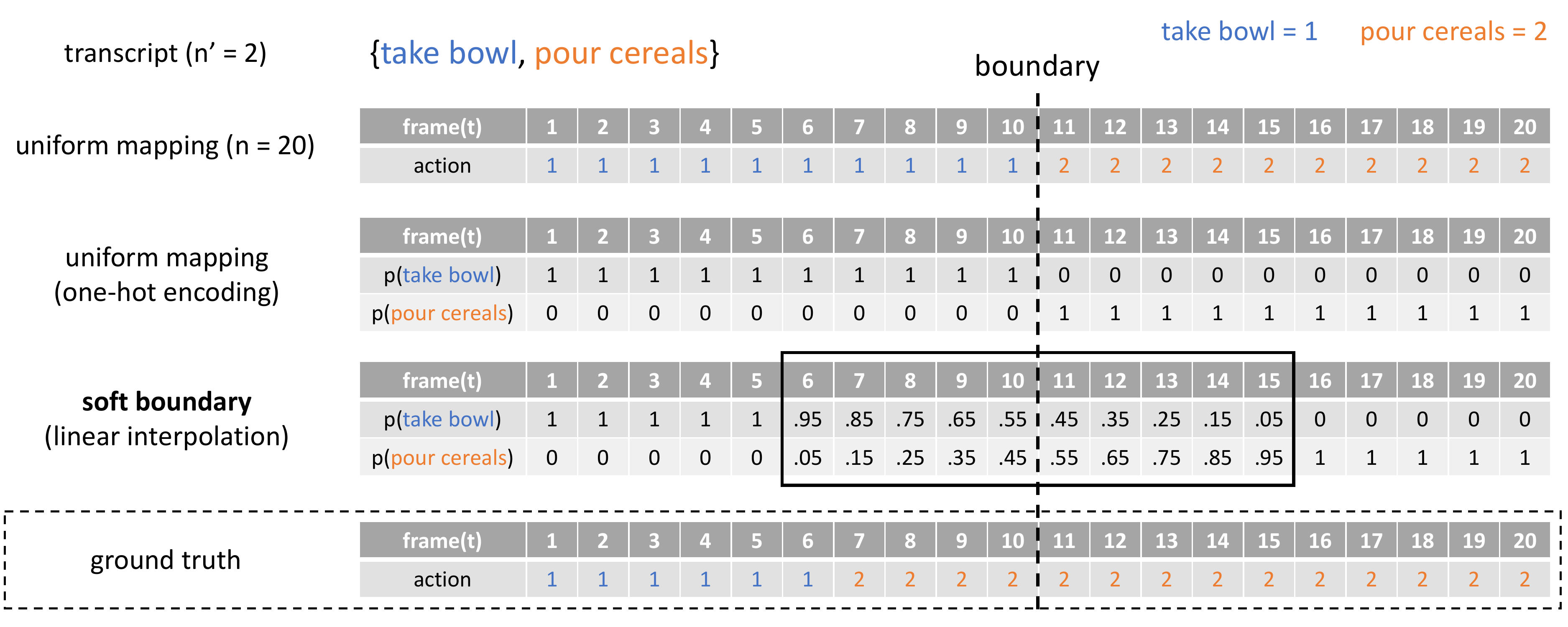}
	\end{center}
	\caption{Example case of the proposed soft boundary mechanism. By using linear interpolation of probabilities, the temporal boundary between two different actions becomes coarse, which makes the generated target more robust and reliable.} 
	\label{soft}
\end{figure}

\begin{figure}[t]
	\begin{center}
		\includegraphics[width=\linewidth]{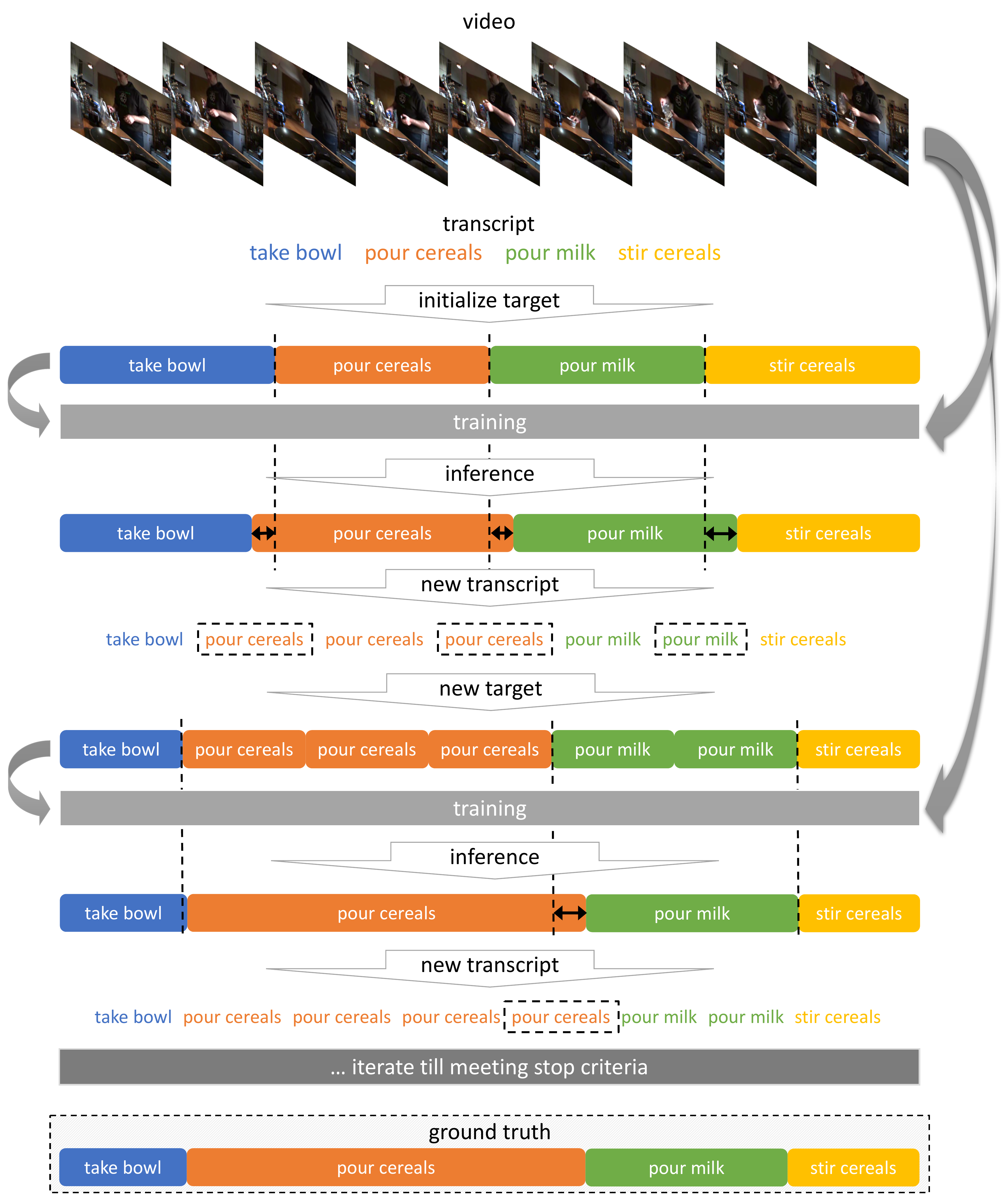}
	\end{center}
	\caption{Overview of the proposed Iterative Soft Boundary Assignment (ISBA) strategy. Under weak supervision, we start from the action transcript and initialize the training target by uniform mapping. A temporal segmentation model is then trained and with which we infer the result on the training set. The action transcript is then refined by the training result and the next iteration starts. Finally until meeting the stop criteria, the final model is used to infer on the training set for the alignment task, and predict on the testing set for the segmentation task. }
	\label{overview}
\end{figure}

In this section, we introduce the overall procedure of our weakly supervised learning mechanism, named Iterative Soft Boundary Assignment (ISBA), which mainly consists of two parts, target generation from transcript and iterative transcript refinement. An overview of the whole procedure is shown in Fig.~\ref{overview}. 

Given a video of \textit{making cereals} as an example, in the training process, we are given the ground-truth action transcript, \ie \{\textit{take\_bowl, pour\_cereals, pour\_milk, stir\_cereals}\}. Notice that, we have only access to the labels and their orders, but not their temporal boundaries in a video. The goal is to localize each action unit in the training video with the given transcript (weakly-supervised action alignment), as well as to predict the actual frame-wise labels for unseen testing videos (weakly-supervised action segmentation).

\subsection{Target Generation with Soft Boundary Assignment}

We start with a linear mapping from the action transcript to video frames. Given a video of length $n$ and its action transcript of length $n'$, usually the linear mapping assign each action label in the transcript $\{A_1, A_2, ... , A_{n'}\}$ to its corresponding $n/n'$ frames as a hard assignment and thus form a target with length $n$. However, this is not an ideal setting for weakly-supervised tasks because such mapping may fail to serve as a good target as the actual lengths of actions vary. In order to encounter this problem, we propose a soft boundary assignment mechanism to set the target. The soft boundary between different actions is defined by a temporal linear interpolation of the probabilities of two labels. As shown in Fig.~\ref{soft}, after uniform initialization with linear mapping and soft boundary assignment, the current target is a sequence with mixed probabilities that generated from linear interpolation. That is to say, we set a \textit{coarse} boundary between different action units to let frames that are close to the boundary have mixed probabilities to have either label. 

One significant advantage of the proposed soft boundary assignment is that, during training, we guide the model to learn discriminative patterns of an action in a smaller temporal interval that is more likely to have the same real ground-truth action label. It works similar to a weighted loss that does not care too much about the boundary, but more specifically let the model choose from the two consecutive actions and thus preserve the ordering information.

We note that, although our change is simple, it is effective. During experiments, we found such setting usually gives a bonus on performance than simply up-sample transcript to video length by hard assignment. A detailed ablation study is performed in Sec.~\ref{abla}. We have also tried other scaling methods such as cubic interpolation, but since the sum of probability should be equal to one as being consistent with the softmax output, and all the probabilities should be greater than zero, the generated target is thus similar to linear interpolation after such normalization.

\subsection{Transcript Refinement with Iterative Training and Inference}

\begin{figure}[t]
	\begin{center}
		\includegraphics[width=\linewidth]{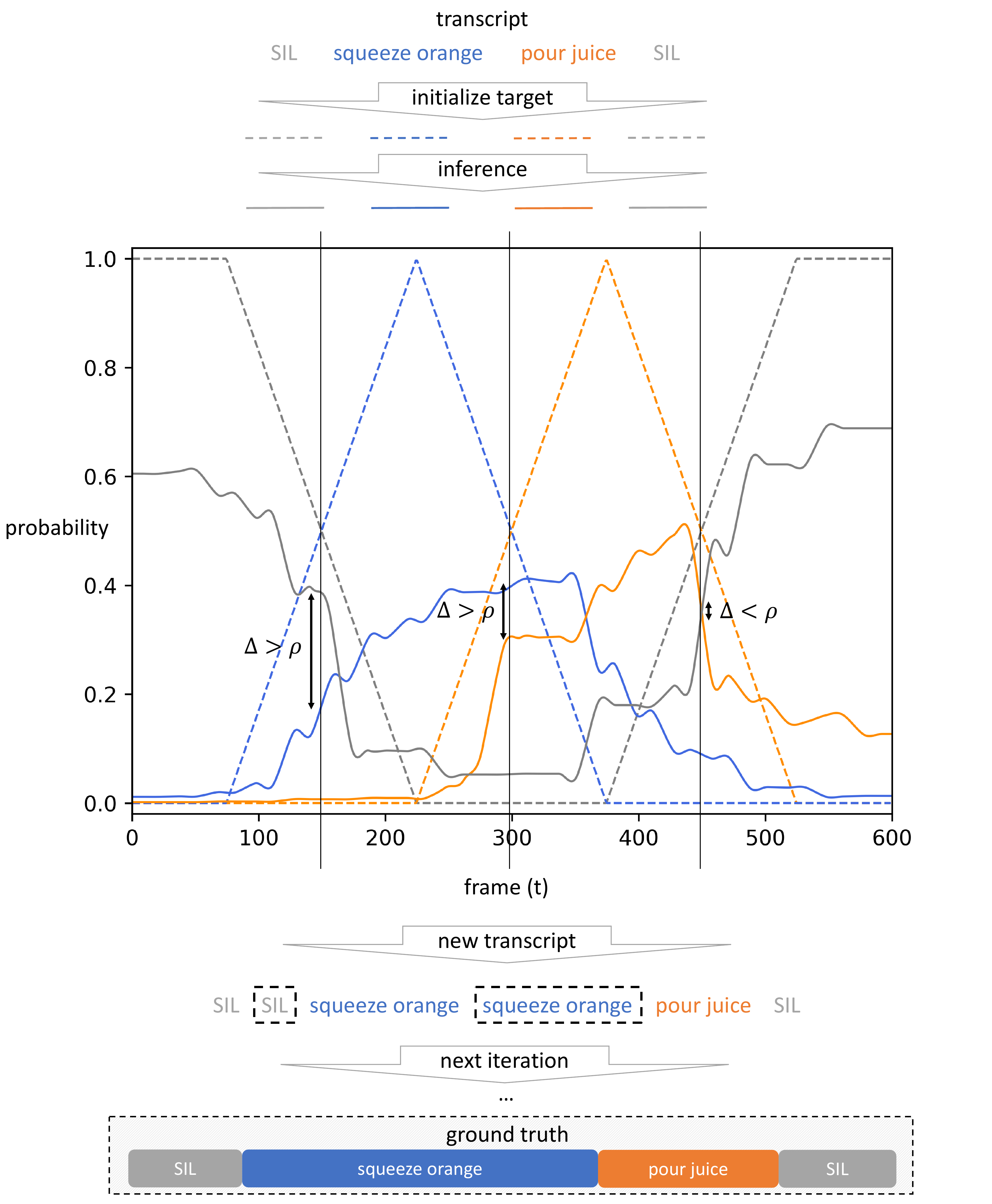}
	\end{center}
	\caption{Example case of the proposed transcript refinement method. Given the initial transcript \{\textit{SIL, squeeze orange, pour juice, SIL}\}, we first generate the target with soft boundary, train the model, and infer the probability. At each temporal boundary between different actions, \ie $t = 150,300,450$, if $\Delta > \rho$ where $\Delta$ is the difference of probabilities and $\rho$ is a threshold, we then insert the action with higher probability into the transcript at corresponding location. In this case, we insert two actions and the new transcript becomes \{\textit{SIL, SIL, squeeze orange, squeeze orange, pour juice, SIL}\}. As a result, the transcript is gradually being refined to mimic the unknown ground truth.}
	\label{prob}
\end{figure}

After each training procedure, a probability sequence is calculated by running inference of the current model on each training sample. As shown in Fig.~\ref{prob}, the predicted probability is likely to be different from the current target. We use an insertion strategy to refine each transcript by utilizing the information learned by the currently-trained model. This operation pushes each action unit to be closer to its possible ground-truth position and its possible length, while preserving their orderings as the weak but exact ground-truth. 

Specifically, for a training video of length $n$ and its action transcript of length $n'$, \ie $\{A_1, A_2, ... ,A_{n'}\}$, at each action boundary $i \in [1, n'-1]$ between different action units, frame $t = (n/n')\cdot i$ is the temporal boundary between action $A_i$ and action $A_{(i+1)}$. If $A_i \neq A_{(i+1)}$, we then observe the inferred probability of both action classes, $P_{A_i}(t)$ and $P_{A_{(i+1)}}(t)$, predicted by the current model at that frame. By setting up a threshold $\rho \in (0,1)$, if: 
\begin{align}
|P_{A_i}(t) - P_{A_{(i+1)}}(t)| > \rho
\enspace,
\end{align}
we insert an action label chosen from $\{A_i,A_{(i+1)}\}$ with the higher probability into the action transcript at a location corresponds to that boundary. Suppose $P_{A_i}(t) > P_{A_{(i+1)}}(t)$, then $A_i$ is the one to insert and the transcript now becomes $\{... , A_i, \pmb{A_i}, A_{(i+1)}, ...\}$. Alg.~\ref{iter} describes the whole iterative process in detail. The runtime of this algorithm is almost linear to different actions in the whole dataset, which is the size of all original transcripts. It is very fast with the simple insertion operation.

Besides, in order to make the process more robust, we also include a randomness parameter $\theta\in(0,0.5]$, according to which the label is chosen from a Bernoulli distribution of the two labels with $p = \theta$. In this case, the probability of insert $A_i$ into transcript is $(1-\theta)$. This idea is inspired from taking random steps used in deep reinforcement learning. Finally, after using the above method to refine all the training transcripts, the next training iteration starts from the new target generation process with the refined transcript. 

The intuition behind this transcript refinement strategy is to iteratively transform the initial action transcript into the \textit{pesudo-supervision}. It can adjust the length of each action unit during every iteration while preserving the ground-truth action orderings. Throughout the refinement, the action transcript becomes less coarse and thus the next target generated from transcript will have \textit{stronger} supervision for training. The most ideal case is that the transcript will get to be more accurate overtime, and at the same time, the training model during each iteration becomes better as the target is more accurate, thus can infer an even better transcript.

\subsection{Stop Criteria}

A proper stop criteria is essential to our approach, as it is very likely to gradually make the refined training target overfit the behavior of the network, which might not really explain the data, and thus miss to fit the unknown ground truth. Also since ISBA itself features a converging behavior (see Alg.~\ref{iter}), other methods such as threshold on frame change~\cite{Richard_2017_CVPR} are not ideal for this case because we can always meet that threshold after a certain number of iterations. In this case, we propose another performance monitoring method to solve this problem. 

\begin{algorithm}
\caption{Transcript Refinement}
\label{iter}
\begin{algorithmic}
\STATE Given $transcript$ in training set
	\FOR{each $action$ in length($transcript$)}
    	\STATE $t = boundary\_frame$
		\IF{($action \neq next\_action$)}
			\IF {$P_{t}(action) - P_{t}(next\_action) > \rho$}
				\STATE insert $action$ into $transcript$ at current location
			\ELSIF {$P_{t}(next\_action) - P_{t}(action) > \rho$}
				\STATE insert $next_action$ into $transcript$ at current location
			\ENDIF
		\ENDIF
	\ENDFOR
\end{algorithmic}
\end{algorithm}

Besides the ordering of actions, what we can also obtain under weak supervision is the occurrence of actions in each video. We utilize this as a measurement to evaluate how the model has learned to recognize actions on video-level. After training and inference, for each training video, we use a global max-pooling through time to get the maximal probability of each actions in video. We then propose a video-level recognition loss, calculated as the binary cross-entropy loss against the ground-truth occurrence of actions, which can be obtained from the action transcript. 

Concretely, given a video of length $n$ and the action set with $k$ classes, the inference result $P$ at  frame-level has dimension $(n,k)$. The global max-pooling shrinks $P$ into $P'$ with dimension $(1,k)$ as the maximal probability of each action class throughout the whole video. Given the ground-truth action occurrence for the same sequence as $Y$, the binary cross entropy loss is calculated as: 
\begin{align}
\mathcal{L}_\text{video} = \sum_{i=1}^{k}[Y_i\log(P'_i) + (1-Y_i)\log(1-P'_i)]
\enspace.
\end{align}
The final loss is averaged over all the samples. In this work, we stop training if this recognition loss does not decrease for three iterations. We then choose the result at the iteration with minimal video-level loss as the final result for a video.

There are several reasons for using such validation. First, it can measure how the model learns to recognize the actions in the video, despite the ordering. It is essential that the model should first recognize the action in order to give the correct ordering and temporal location. Second, its real ground truth can be obtained from transcript under the weakly-supervised setting. Third and most importantly, since we do not use that loss directly for back propagation during training, it is less likely that the model will overfit that loss. Given the fact that TCFPN does classification on each frame, the video-level context is harder to be learned from frame-level loss. 

Generally speaking, we use this sequence recognition loss to find out the best condition that the model learns to model discriminative patterns of different actions during the whole iterative training process, although the training targets are generated from different transcripts. 

\begin{figure}[t]
	\begin{center}
		\includegraphics[width=\linewidth]{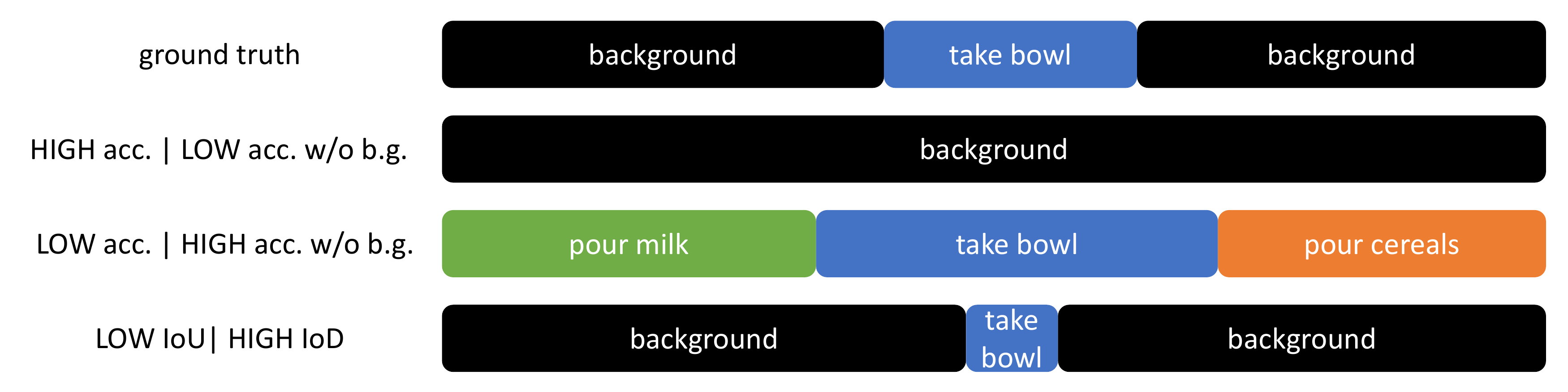}
	\end{center}
	\caption{Comparison among different metrics.}
	\label{metric}
\end{figure}

\section{Experiments}
\label{sec:exp}

In this section, we describe in detail the experimental results. We experiment with three different tasks, e.g., fully-supervised action segmentation, weakly-supervised action segmentation, and weakly-supervised action alignment, on two datasets, e.g., Breakfast~\cite{kuehne2014language} and Hollywood Extended~\cite{bojanowski2014weakly}. The Breakfast dataset has over 1.7k video sequences of cooking in the kitchen environment with an overall duration of 66.7h. Each video is labeled with a set of 48 action classes with a mean of 4.9 instances per video. The Hollywood Extended dataset contains 937 video sequences from different Hollywood movies. It features a set of 16 different action classes overall and a mean of 2.5 action instances per video. Surpassing \cite{ni2014multiple,stein2013combining}, these two datasets are among the largest datasets for evaluating action segmentation. 

We use four different metrics to evaluate the performance, making it more general and easy to compare with previous and future work on this topic. Besides, we also do an ablation study to further explore the proposed method.

\subsection{Metrics}

Performance metric is very task- or need-specific for video understanding problems. Most of other works propose or stick to only one or two specific metrics. In this work, we use four different metrics to evaluate qualitative results on all three tasks. Examples of comparison among different metrics are shown in Fig.~\ref{metric}. 

Frame-wise accuracy (\textbf{Acc.}) is a common metric for action segmentation, which directly evaluates how many frames are correctly labeled. A drawback of this metric is that, if frames in a are dominated by a single class, \eg \textit{background}, one can achieve high accuracy by simply classifying everything as that class. 
To overcome the limitations, we propose another metric, frame-wise accuracy without background (\textbf{Acc.$-$b.g.}), which computes the accuracy for all frames except background frames. In this case, the metric emphasizes on evaluating frames with a real action label. 

Another set of metrics are Jaccard measures, which observe how the prediction and ground truth overlaps each other. In this work, we obtain results of both intersection over union (\textbf{IoU}) and intersection over detection (\textbf{IoD}). Given a ground-truth action interval $I^*$ and a prediction interval $I$, IoU is measured as $|I \cap I^*| / |I \cup I^*|$ and IoD is measured as $|I \cap I^*| / |I| $. IoU requires accurate temporal segmentation to achieve high score, while IoD requires the prediction to be included in the ground truth. Thus, IoD is more related to the detection task instead of action segmentation. Since previous work~\cite{huang2016connectionist, bojanowski2014weakly, Richard_2017_CVPR} tend to use IoD as a measure, we also include the IoD metric for comparison although it is not so relevant to the segmentation task. 

\begin{table}[t]
	\caption{Results of \textbf{fully-supervised action segmentation}. The proposed network TCFPN outperforms state-of-the-art methods on both datasets with all four metrics. (*result obtained from the author of \cite{Richard_2017_CVPR}; **our implementation)} 
	\centering
	\vspace{2mm}
	\label{sup-seg}
	\begin{tabularx}{\linewidth}{X|rccc}
		\toprule
		\textbf{Breakfast} &  Acc. &  Acc.$-$b.g.& IoU & IoD\\
		\midrule
		HTK(64)~\cite{kuehne2016end} & 56.3 &-&-&-\\
		HTK~\cite{kuehne2017weakly}                 & 50.7 &-& 36.1&-   \\
		GRU~\cite{Richard_2017_CVPR}* & \textbf{60.6} & - & - & - \\
		ED-TCN~\cite{Lea_2017_CVPR}**         & 43.3 & 43.6& 29.4&42.0   \\
		\midrule
		TCFPN  & 52.0 &\textbf{52.6} & \textbf{36.7} &\textbf{54.9} \\
		\bottomrule
	\end{tabularx}
	\vspace{1mm}  
	
	\begin{tabularx}{\linewidth}{X|rccc}
		\toprule
		\textbf{Hollywood Ext.} &  Acc. &  Acc.$-$b.g. & IoU & IoD\\
		\midrule
		HTK~\cite{kuehne2017weakly}                  & 39.5 &- &8.4 & -  \\
		ED-TCN~\cite{Lea_2017_CVPR}**          & 36.7 & 27.3 & 10.9 & 13.1  \\
		\midrule
		TCFPN  & \textbf{54.8} & \textbf{33.1} & \textbf{20.4} & \textbf{28.8} \\
		\bottomrule
	\end{tabularx}
\end{table}

\subsection{Implementation Details}

In order to make the results comparable, for both datasets we obtain features used in previous work~\cite{kuehne2017weakly, kuehne2016end} from the authors, which are computed using improved dense trajectories (iDT) and Fisher vectors (FV) with PCA and GMM, as described in~\cite{kuehne2016end}. Frames are down-sampled to about 1 fps. 
We will release our implementations along with all parameters upon acceptance to facilitate future research.

\subsection{Comparing to State-of-the-Art Methods}

We compare the proposed methods under three different tasks to the state-of-the-art. Since other work usually only use one or two metrics, we leave the blank for those missing evaluations.

\vspace{1mm}
\noindent \textbf{Fully-Supervised Action Segmentation}

\noindent We first evaluate the proposed temporal segmentation network TCFPN in fully-supervised setting. We also implement the ED-TCN~\cite{Lea_2017_CVPR} as our baseline method, which has not been implemented on the two datasets yet. As shown in Table~\ref{sup-seg}, TCFPN outperforms ED-TCN to a large extent and shows competitive results to state-of-the-art methods. The HTK(64)~\cite{kuehne2016end} uses the same HTK system as \cite{kuehne2017weakly}, but a better feature with GMM size of 64. Besides, the proposed method does not contain recurrent connections, e.g., LSTMs, and thus the training can be very efficient with highly-parallelized computation. 

\vspace{1mm}
\noindent \textbf{Weakly-Supervised Action Segmentation}

\begin{table}[t]
	\caption{Results of \textbf{weakly-supervised action segmentation}. The proposed TCFPN + ISBA outperforms other state-of-the-art methods on both datasets with most of the metrics. (*from~\cite{huang2016connectionist})}
	\vspace{2mm}
	\centering
	\label{weak-seg}
	\begin{tabularx}{\linewidth}{X|rccc}
		\toprule
		\textbf{Breakfast} &  Acc. &  Acc.$-$b.g. & IoU & IoD\\
		\midrule
		OCDC~\cite{bojanowski2014weakly}*                & 8.9 &- &-&-      \\
		HTK~\cite{kuehne2017weakly}                  & 25.9 &-&9.8&-   \\
		ECTC~\cite{huang2016connectionist}                & 27.7&-&-&-      \\
		GRU reest.~\cite{Richard_2017_CVPR}    & 33.3 &-&-&-     \\
		\midrule
		ED-TCN + ISBA          & 32.0 & 28.8  & 18.4 & 30.6  \\
		TCFPN + ISBA  & \textbf{38.4} & \textbf{38.4} & \textbf{24.2}  & \textbf{40.6} \\
		\bottomrule
	\end{tabularx}
\vspace{1mm}

	\begin{tabularx}{\linewidth}{X|rccc}
		\toprule
		\textbf{Hollywood Ext.} &  Acc. &  Acc.$-$b.g. & IoU & IoD\\
		\midrule
		HTK~\cite{kuehne2017weakly}                  & \textbf{33.0} &-&8.6  &- \\
		GRU reest.~\cite{Richard_2017_CVPR}    & - &- &11.9 &-  \\
		\midrule
		ED-TCN + ISBA        & 27.8 &29.4  & 8.9  & 10.7 \\
		TCFPN + ISBA  & 28.7  &  \textbf{34.5 }& \textbf{12.6} & \textbf{18.3} \\
		\bottomrule
	\end{tabularx}
\end{table}

\noindent We evaluate the proposed ISBA strategy with both ED-TCN and TCFPN on the weakly-supervised action segmentation, which is the main task of this work. As shown in Table~\ref{weak-seg}, the proposed TCFPN+ISBA outperform other state-of-the-art methods to a large extent on almost every metric. The only exception is for the frame-wise accuracy on Hollywood Extended dataset, which is due to a large amount of \texttt{background} frames. Although the HTK~\cite{kuehne2017weakly} has a better frame-wise accuracy, its IoU is much lower than other state-of-the-art methods and our proposed method. Our method outperforms HTK significantly on Breakfast dataset.

Another observation, when jointly considering the result from previous fully-supervised action segmentation, is that when TCFPN shows competitive or slightly better performance than other methods, TCFPN + ISBA shows much stronger performance on weakly-supervised task. More specifically, for Breakfast dataset, TCFPN is $5.8\%$ worse than GRU on frame-wise accuracy if fully supervised. However when it comes to weakly-supervised task, TCFPN + ISBA is $5.1\%$ better than GRU and HMM with re-estimation~\cite{Richard_2017_CVPR}. Thus, we regard the proposed ISBA mechanism as the main boost in the weakly supervised action segmentation task. 

\vspace{1mm}
\noindent \textbf{Weakly-Supervised Action Alignment}

\begin{table}[t]
	\caption{Results of \textbf{weakly-supervised action alignment} on training set. The proposed TCFPN + ISBA outperforms other state-of-the-art methods on both datasets with most of the metrics. (*from author's plot)}
	\vspace{2mm}
	\centering
	\label{weak-al}
	\begin{tabularx}{\linewidth}{X|rccc}
		\toprule
		\textbf{Breakfast} &  Acc. &  Acc.$-$b.g. & IoU & IoD\\
		\midrule
		OCDC~\cite{bojanowski2014weakly}                & - &- &-&23.4      \\
		HTK~\cite{kuehne2017weakly}                  & 43.9 &-&26.6&42.6  \\
		ECTC~\cite{huang2016connectionist}*                & $\sim$35 &-&-&$\sim$45     \\
		GRU reest.~\cite{Richard_2017_CVPR}    & - &-&-&47.3     \\
		\midrule
		ED-TCN + ISBA     & 52.7 &50.5  & 33.5 &51.4  \\
		TCFPN + ISBA  & \textbf{53.5} & \textbf{51.7} & \textbf{35.3}  & \textbf{52.3}\\
		\bottomrule
	\end{tabularx}
\vspace{1mm}

	\begin{tabularx}{\linewidth}{X|rccc}
		\toprule
		\textbf{Hollywood Ext.} &  Acc. &  Acc.$-$b.g. & IoU & IoD\\
		\midrule
		OCDC~\cite{bojanowski2014weakly}  & -&-&-& 43.9 \\
		HTK~\cite{kuehne2017weakly}                  & 49.4 &-&29.1  & \textbf{46.9} \\
		ECTC~\cite{huang2016connectionist}*  & -&-&-&  $\sim$41 \\
		GRU reest.~\cite{Richard_2017_CVPR}    & - &- & - & 46.3 \\
		\midrule
		ED-TCN + ISBA    & 50.3 & 32.4  & 26.2  & 34.8 \\
		TCFPN + ISBA  & \textbf{57.4} &  \textbf{36.1} & 22.3 & 39.6 \\
		\bottomrule
	\end{tabularx}
\end{table}

\noindent We also show the weakly-supervised action alignment task, which aims to align the given transcript to its proper temporal location. This task is in parallel with segmentation because we iteratively refine the transcript and thus have a better segmentation model with improved training target. We report the alignment result on the testing set by iteratively refine the testing scripts for $10$ iterations with the best model determined by stop criteria. 

As shown in Table~\ref{weak-al}, the proposed TCFPN+ISBA again shows superior result on Breakfast dataset. For Hollywood Extended dataset, many works choose to use IoD as the only metric to evaluate the detection performance. In this work, we assume the video is well labeled with action transcript and there are less background or meaningless labels. As a result, we achieve a better IoU but a lower IoD, because the network is less likely to label the majority of frames as background.

\subsection{Analysis and Ablation Study}\label{abla}

In this section, several experiments are done to further evaluate the proposed method. First, we evaluate the speed of the proposed method. Second, we investigate the effectiveness of the proposed soft boundary mechanism. Third, we visualize detailed result of an example training process with all the iterations included, showing how the method learns from weak labels and how the proposed stop criteria works. Experiments give more direct insights about  performance and capability of the proposed method.

\noindent \textbf{Speed.} \quad On Breakfast dataset, TCFPN+ISBA takes only 4 minutes for one iteration of training including inference and prediction on the training set, averagely around half an hour for the whole training process. Comparing to previous work, OCDC~\cite{bojanowski2014weakly} takes about 2 hours to train; HTK~\cite{kuehne2017weakly} takes 23.7 minutes just for segmentation prediction.

\noindent \textbf{Soft Boundary.} \quad In addition to experiments described in the previous section, we train the TCFPN+ISBA model without soft boundary on Breakfast dataset. In this case, the transcript is simply repeatedly up-sampled to the video length to generate the training target. The comparison result is shown in Table~\ref{tbsoft}. By adding the soft boundary mechanism, the final results gain improvement on most of the metrics. Although the margin is small, the improvement is consistent and can be easily applied to other weakly-supervised systems. Another benefit from soft boundary is that ISBA converges more quickly.

Overall, the soft boundary mechanism stands for the idea of using a mixture of probabilities to represent the boundary between two different actions. We believe such representation and mechanism are important in the field of action modeling, where you can not always tell the exact changing point of an action, and thus worths further exploration.

\begin{table}[t]
	\caption{\textbf{Ablation study} of weakly-supervised action segmentation and alignment on Breakfast dataset. By adding the soft boundary mechanism, the final results gain improvement on most of the metrics.}
	\vspace{2mm}
	\centering
	\label{tbsoft}
	\begin{tabularx}{\linewidth}{X|rccc}
		\toprule
		Breakfast (seg.)  &  Acc. &  Acc.$-$b.g. & IoU & IoD\\
		\midrule
		TCFPN + ISBA  & \textbf{38.4} & \textbf{38.4} & \textbf{24.2}  & 40.6 \\
		Above w/o soft b.d. & 37.8 & 38.1 & 24.1 & \textbf{41.8} \\
		\bottomrule
	\end{tabularx}
\vspace{1mm}

	\begin{tabularx}{\linewidth}{X|rccc}
	\toprule
	Breakfast (align.)  &  Acc. &  Acc.$-$b.g. & IoU & IoD\\
	\midrule
	TCFPN + ISBA  & \textbf{56.7} & \textbf{55.9} & \textbf{38.7}  & \textbf{54.0} \\
	Above w/o soft b.d. &  55.7 & 55.8 & 38.1 & 53.0 \\
	\bottomrule
	\end{tabularx}
\end{table}

\begin{figure}[t]
	\begin{center}
		\includegraphics[width=\linewidth]{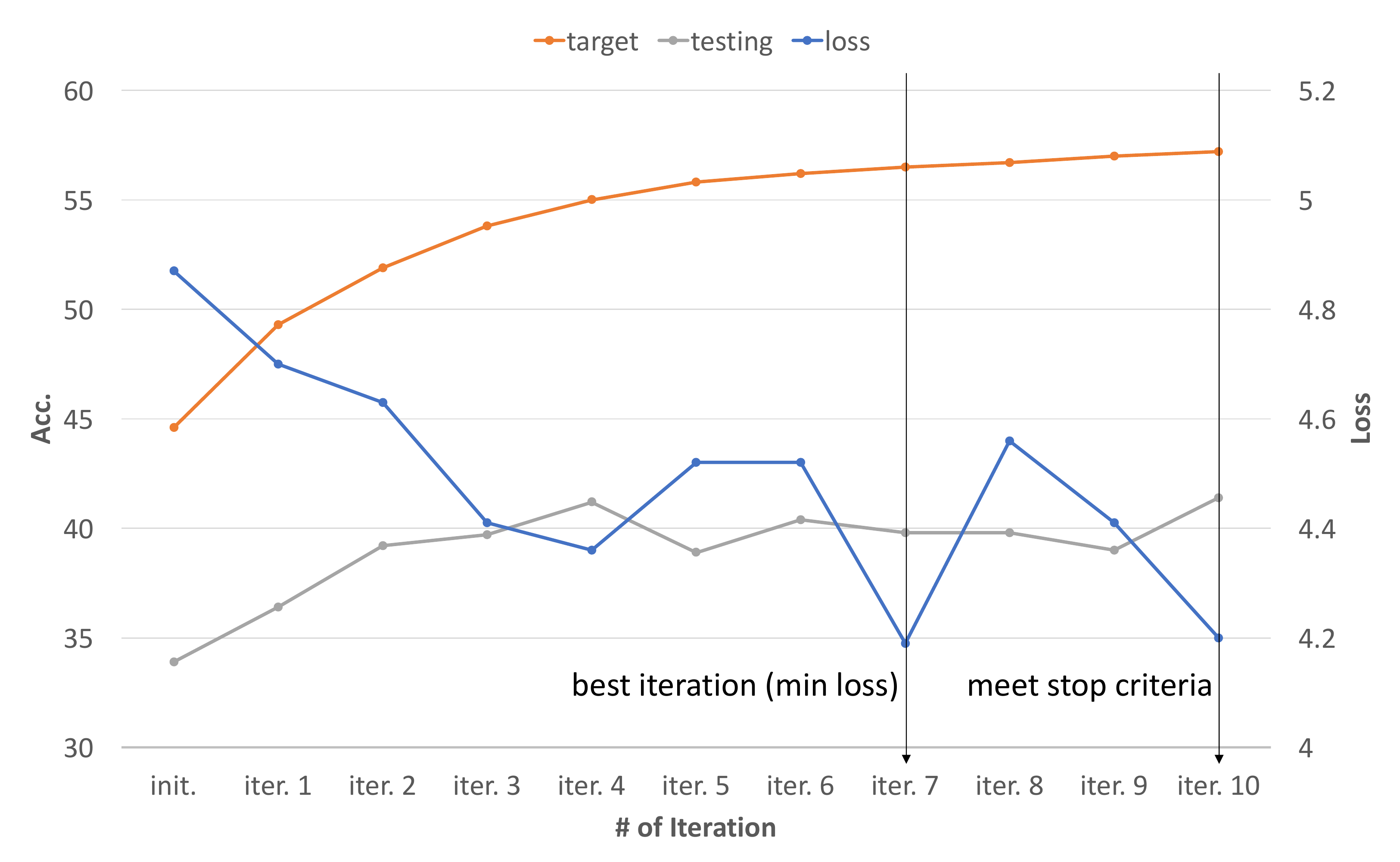}
	\end{center}
	\caption{Example training process on the first split of Breakfast dataset. Stop criteria is met at iteration 10 as the loss does not decrease for three iterations from iteration 7, which is the best iteration and used as the final result.}
	\label{stop1}
\end{figure}

\begin{figure}[t]
	\begin{center}
		\includegraphics[width=\linewidth]{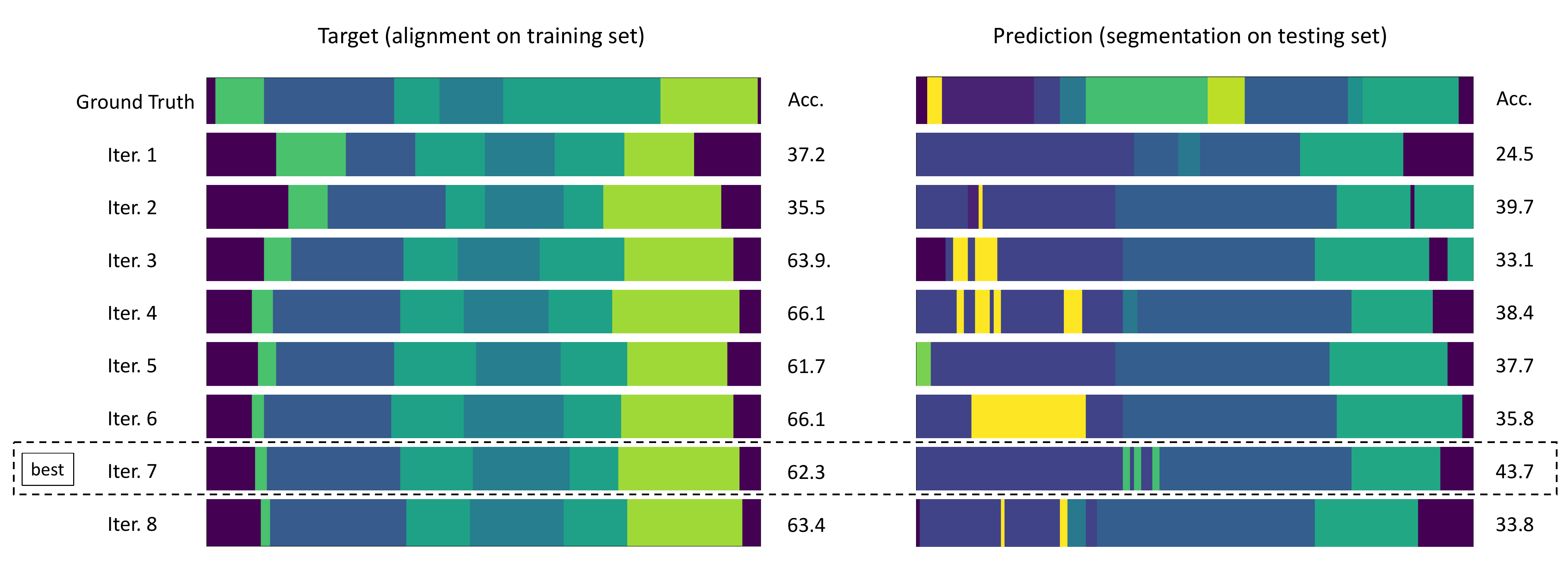}
	\end{center}
	\caption{Two example videos in training and testing set of Breakfast dataset during the iterative refinement. On the left-hand-side, the training target starts from a unique mapping and gets refined over time. Note that the targets are probabilities generated with soft boundary, and visualization shows only the label with largest probability. On the right-hand-side, the testing sample is predicted by models trained during each iteration.}
	\label{stop2}
    \vspace{-2mm}
\end{figure}

\noindent \textbf{Stop Criteria.} \quad We visualize the whole training process on the first split of Breakfast dataset. Fig.~\ref{stop1} shows the frame-wise accuracy for both alignment task on training set and segmentation task on testing set, together with the recognition loss used as our stop criteria. One thing to mention is that we can only access the video-level recognition loss during training, among these three values. By using the proposed stop criteria, we can quickly find the best recognition model without extra training. 

As we see, although the alignment accuracy keeps being improved, the testing accuracy begins to converge after a few iterations. The proposed recognition loss aligns well against the actual testing accuracy. We also do additional experiment using the stop criteria from~\cite{Richard_2017_CVPR} on the first split of Breakfast dataset, resulting $21$ iterations with $38.7$ Acc., which means almost twice iterations with lower performance, comparing to the $10$ iterations with $39.8$ Acc. from Fig.~\ref{stop1}.

Fig.~\ref{stop2} shows example results on training and testing set, respectively. As the training target gradually being refined to its proper length and location, the testing result also gets better as the model becomes more precise. The proposed method is designed to find useful information throughout the whole dataset in an iterative fashion. 

\section{Conclusion}

In this work, we propose ISBA as a novel strategy for weakly-supervised action segmentation and alignment. It features an iterative training procedure with transcript refinement and soft boundary assignment, together with a video-level loss metric proposed as the stop criteria. We also propose TCFPN, a new temporal convolutional network for supervised action segmentation, which can be fast-trained and shows competitive performance to state-of-the-art methods. The whole system TCFPN+ISBA outperforms state-of-the-art on both weakly-supervised action segmentation and alignment. Our training strategy is general and can be integrated with other work or used in other tasks, to facilitate future research in video action understanding.


\vspace{1em}

\noindent {\textbf{Acknowledgement.} \quad This work was supported by NSF BIGDATA 1741472. We gratefully acknowledge the gift donations from Markable, Tencent and the support from NVIDIA with the donation of GPUs. This article solely reflects the opinions and conclusions of its authors and not the funding agents.}

{\small
\bibliographystyle{ieee}
\bibliography{egbib}
}

\end{document}